\documentclass[10pt]{article}
\usepackage[margin=1in]{geometry}
\usepackage[T1]{fontenc}
\usepackage[utf8]{inputenc}
\usepackage{graphicx}
\usepackage{booktabs}
\usepackage{amsmath}
\usepackage{amssymb}
\usepackage{multirow}
\usepackage[numbers]{natbib}
\usepackage[hidelinks]{hyperref}
\usepackage{xcolor}

\newcommand{\dW}{\Delta W}

\title{Learning Only What Valid Adapters Can Express:\\
Subspace-Constrained Adaptation Against Fine-Tuning Poisoning}
\author{Fabien Polly\thanks{Independent researcher. Code and data:
\url{https://github.com/infinition/z-manifold}}}
\date{July 2026}

\begin{document}
\maketitle

\begin{abstract}
Parameter-efficient fine-tuning still leaves a broad space of behavior-changing
updates reachable, so a poisoned objective can be represented and optimized. We
study an alternative: adaptation constrained to the subspace estimated from a
trusted pool of existing task adapters. On flan-t5-large with 196 public LoRA adapters, we show that (1) the
functionally relevant content of an adapter lies in a low-dimensional shared
subspace, 30 to 38 percent of its weight norm being redundant under the
evaluated task distributions;
(2) gradient adaptation restricted to 128 coordinates on this subspace matches
full LoRA fine-tuning on clean classification data, while under targeted label
inversion LoRA collapses to 3--26 percent exact match and the constrained
learner keeps 62--96 percent on the tasks the pool covers; (3) the constrained
learner cannot fit corrupted data, its adaptation loss separating clean from
garbage by two orders of magnitude ($120\times$), an out-of-distribution signal
for free; and (4) against an adaptive backdoor attacker who optimizes within the
subspace, the attack is blocked (8 percent success versus 100 for LoRA) on the
task where its target behavior is unlike anything in the pool, and only partially
blocked (85 percent) when the target coincides with a common pool behavior. On
these two tasks the outcome is consistent with how close the target is to the
pool's directions, which suggests but does not establish a pool-relative
boundary. The mechanism trades peak
plasticity for these properties: on tasks the pool covers poorly, unconstrained
fine-tuning
wins, and the guarantee assumes the pool itself is trusted. Code and data are
public.
\end{abstract}

\section{Introduction}
Fine-tuning is an attack surface. A consumer medical assistant maliciously
personalized to recommend a dangerous dose, a corporate model taught by a
poisoned document to exfiltrate its database on a trigger word, a brand chatbot
absorbing toxic user chats, a model fine-tuned on scraped web text that a
coordinated edit campaign has seeded with falsehoods: every adaptation on data
the operator does not fully control is a vulnerability. A few hundred corrupted
instructions suffice to compromise an aligned model~\citep{wan2023poisoning,
qi2024finetuning}. Existing defenses inspect the data (filtering, influence
functions) or dampen the update (regularization); all are heuristic in the
sense that a successful attack remains expressible, the defense just makes it
harder to reach.

We study a defense that shrinks expressibility instead. The weight updates a
model acquires when learning legitimate tasks do not fill update space; they
concentrate in a low-dimensional shared subspace. If adaptation is constrained
to that subspace, the set of reachable updates is greatly reduced, and we find
empirically that the poisoned objectives in our threat models require updates
outside it: they cannot be fit, and that failure is visible in the training
loss. We are careful not to overclaim: a subspace spanned by legitimate
adapters can still contain far extrapolations and combinations unlike any
single adapter, so this is empirical protection against the attacks we test,
not a proof that no harmful behavior is expressible (Section~\ref{sec:limits}).

We instantiate the idea with the affine span of a public pool of LoRA adapters,
and evaluate it on a real language model against an equal-data,
equal-optimization-step LoRA baseline. Our contributions: (1) an analysis
of the functional geometry of 196 public adapters showing that 30 to 38
percent of an adapter's weight norm is functionally redundant under the
evaluated distributions; (2) a constrained adaptation mechanism with 128 trainable coordinates
that matches LoRA fine-tuning on clean data; (3) a poisoning study showing an
order-of-magnitude robustness gap under targeted label inversion; (4) a
built-in out-of-distribution signal separating clean from garbage adaptation
data by two orders of magnitude. Everything runs on one consumer GPU and all
code, data, and seeds are public.

\section{Related work}
\textbf{Latent adaptation.} LEO~\citep{rusu2019leo} performs gradient descent
on a latent code with a frozen decoder in its inner loop, for few-shot
classification, decoding only a final linear layer; its headline pipeline adds
weight-space fine-tuning after decoding, and no safety property is evaluated.
Our test-time mechanism is LEO's ablated z-only variant, applied to the LoRA
deltas of a language model and evaluated for safety.
\textbf{Feed-forward adapter generation.} Text-to-LoRA~\citep{charakorn2025t2l}
and Drag-and-Drop LLMs~\citep{dnd2025} generate LoRA parameters in a single
forward pass from task descriptions or prompts; neither performs test-time
optimization nor evaluates poisoning, OOD detection, or forgetting.
Text-to-LoRA's appendix documents the weight-space misalignment of
independently trained adapters that we also observe and work around by
operating in $\dW$ space.
\textbf{Adapter subspaces (closest work).} EigenLoRAx~\citep{eigenlorax2025}
recycles existing adapters into a principal subspace and adapts new tasks by
learning only coefficients within it, for efficiency; the mechanism is close to
ours. Compress-then-Serve~\citep{cts2024} learns shared bases over many LoRAs
for serving, and LoraHub~\citep{lorahub2024} composes adapters for cross-task
generalization. Our mechanism (optimize coefficients in an adapter subspace) is
not the contribution; these works already establish that new tasks can be
adapted inside such a subspace. Our contribution is to use the subspace as an
\emph{expressivity barrier against poisoning} and to show that the optimization
residual is itself a rejection signal. In one line: EigenLoRAx is efficiency
and recycling, LoraHub is composition, ours is safety by restriction of the
reachable set plus detection by non-learnability.
\textbf{Weight-space structure.} Model zoos~\citep{schurholt2022} study
populations of trained networks; permutation symmetry~\citep{gitrebasin2023}
explains why independently trained networks are misaligned in raw weight space,
which motivates our use of the gauge-invariant $\dW$ representation.
\textbf{Poisoning of fine-tuning.} Instruction tuning can be poisoned with a
few hundred examples~\citep{wan2023poisoning}, and even benign fine-tuning
degrades safety alignment~\citep{qi2024finetuning}. These works motivate the
threat model; existing mitigations are data-side (filtering, provenance) or
update-side (regularization, constrained norms), all leaving the attack
expressible. We instead make the attack geometrically unreachable.

\section{Method}
\label{sec:method}
Pool of $n$ adapters with deltas $\dW_j = B_j A_j$ (gauge invariant; we never
compare raw $(A,B)$ factors). Frobenius inner products are computed in factored
form, $\langle \dW_i, \dW_j \rangle = \mathrm{tr}\big((B_i^\top B_j)(A_j
A_i^\top)\big)$, giving a Gram matrix $G \in \mathbb{R}^{n \times n}$ without
materializing any $\dW$.

\textbf{Basis.} For a held-out task $i$ we take the other $n{-}1$ adapters,
double-center their Gram submatrix, and compute its eigendecomposition; the
top $K$ eigenvectors give an orthonormal basis of the pool's span, expressed as
coefficient vectors over the adapters. This is a full $(n{-}1)\times(n{-}1)$
eigendecomposition recomputed per held-out task; with $n=196$ it costs a few
milliseconds and is exact, not an approximation or a rank-one update. Because
the pool of task $i$ excludes $i$, the target adapter is never in its own basis,
which prevents leakage in the reconstruction experiments (Section~\ref{sec:func}).

\textbf{Adaptation.} We optimize a code $z \in \mathbb{R}^K$ and set
$\dW(z) = \sum_j g_j(z)\, \dW_j$ with the mixing weights $g(z) = M z + b$ affine
in $z$: $M \in \mathbb{R}^{n \times K}$ carries the basis eigenvectors scaled by
the empirical per-component standard deviation, and $b$ is the pool-mean weight,
so $z=0$ (its initialization) decodes to the mean adapter. Only $z$ is trained
by gradient descent; the base model and the pool are frozen. In this base method
the reachable set $\{\dW(z)\}$ is therefore the affine span of the pool. The
implementation wraps the $q$ and $v$ projections with stacked factor matrices;
the overhead is about $3\times$ a forward pass on the wrapped projections and no
basis is ever materialized. Section~\ref{sec:generator} replaces the affine
$g$ with a nonlinear generator to test whether a curved manifold helps.

\section{Experiments}
Setup: flan-t5-large (783M, bf16), 196 LoraHub adapters (rank 16, $q$ and $v$,
144 modules, 9.4M parameters each), P3 tasks, 128 training examples per task,
identical step budgets for both methods (8 epochs, batch 8), LoRA baseline
freshly initialized with rank 16. Three seeds per cell. A 12 GB consumer GPU
suffices.

\subsection{Structure of the adapter pool}
\label{sec:spectrum}
\begin{figure}[t]
  \centering
  \includegraphics[width=0.55\linewidth]{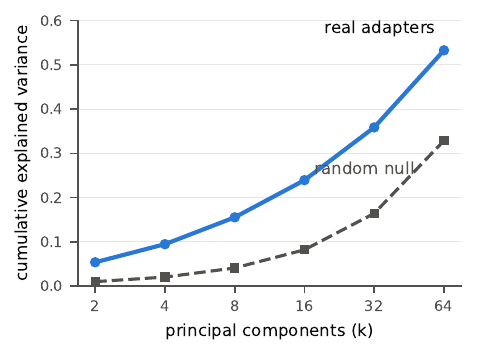}
  \caption{Cumulative explained variance of the 196 adapters in $\dW$ space
  versus a norm-matched random-LoRA null. Structure is real (top-32 captures
  35.9\% versus 16.4\% for the null) but not naively low-dimensional
  (effective dimension 129 of 196). Within-family cosine similarity is 0.204
  versus 0.018 across families.}
  \label{fig:spectrum}
\end{figure}
The effective dimension of 129 out of 196 is itself informative: the safety of
the method does not come from crushing the model into a tiny subspace, but from
removing precisely the directions no legitimate adapter uses. The constrained
learner keeps most of the pool's expressive capacity while excluding the rest.

\subsection{Functional dimension}
\label{sec:func}
\begin{figure}[t]
  \centering
  \includegraphics[width=0.7\linewidth]{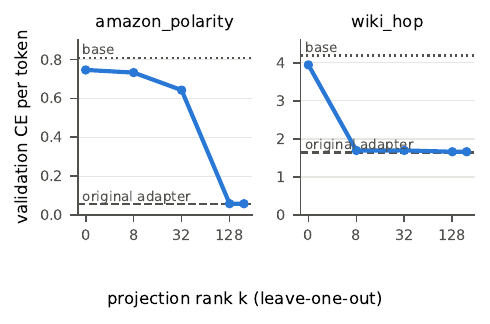}
  \caption{Validation cross-entropy of an adapter reconstructed by
  leave-one-out projection on the top-$k$ principal directions of the other
  195. Function is fully recovered at $k{=}8$ (wiki\_hop) to $k{=}128$
  (amazon\_polarity) while 30 to 38 percent of the weight norm, orthogonal to
  the pool, is discarded with no functional cost.}
  \label{fig:functional}
\end{figure}

The spectrum measures geometry; here we ask how much of an adapter is
functionally necessary. For a held-out adapter, we reconstruct it from the
leave-one-out basis of the other 195: we keep the projection of its $\dW$ onto
the top-$k$ principal directions of the pool, apply that reconstruction to
flan-t5-large, and read the validation cross-entropy per token on the adapter's
own task (Figure~\ref{fig:functional}). We do this on the tasks where the
original adapter measurably beats the base model, so that a change in accuracy
is meaningful. The function is fully recovered at a modest rank, $k=8$ for
wiki\_hop and $k=128$ for amazon\_polarity, even though 30 to 38 percent of the
adapter's weight norm lies outside the pool's span and is thrown away in the
reconstruction. That discarded component is orthogonal to every other adapter
and carries no measurable function on the task; it is the degeneracy of
independently trained low-rank updates, not signal. In other words, the part of
an adapter that matters already lives in the shared subspace, which is what
makes constraining adaptation to that subspace viable rather than crippling.

\subsection{Poisoning}
\begin{figure}[t]
  \centering
  \includegraphics[width=0.55\linewidth]{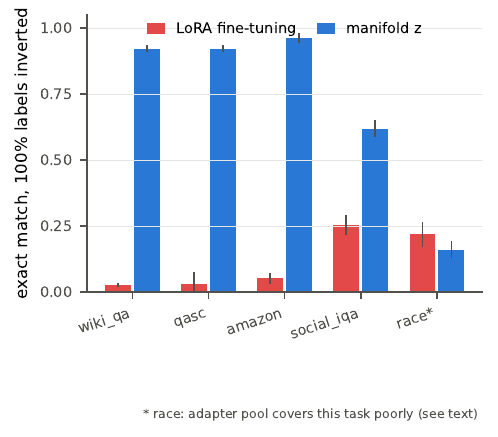}
  \caption{Exact match on clean validation after adapting on 128 examples with
  100 percent targeted label inversion, per task (mean and standard deviation
  over 3 seeds). On the four tasks the pool covers, constrained adaptation
  keeps 0.62 to 0.96 while LoRA collapses to 0.03 to 0.26. race is the
  weak-pool case (Table~\ref{tab:inversion}): both are low and the constraint
  does not help.}
  \label{fig:poison}
\end{figure}
Under intra-task label shuffling (a weak poison preserving the label marginal),
LoRA often resists through benign memorization; under targeted inversion the
gap is an order of magnitude (Figure~\ref{fig:poison},
Table~\ref{tab:inversion}). The prior is capped by pool quality: on tasks
whose pool adapters are weak (race), LoRA reaches lower clean CE than
constrained adaptation. This trade-off is inherent and we report it as such.

\textbf{Why the poison is not expressible.} A label-inverted task requires a
weight update that maps each input to the opposite of its correct answer. No
legitimate adapter in the pool implements such a mapping, so the update has a
large component orthogonal to the pool's span. Constrained adaptation can only
realize the in-span projection of that update, which for a systematic inversion
is close to the pool mean (a weak, non-inverting adapter); the inverting
direction is simply unavailable. LoRA, unconstrained, follows the full update
and learns the inversion faithfully. This is the mechanistic content of the gap
in Table~\ref{tab:inversion}, and it predicts the failure mode we do observe:
the method degrades gracefully toward a generic adapter rather than toward a
harmful one.

\begin{table}[t]
  \centering
  \caption{Exact match on clean validation after adapting on 128 examples,
  mean $\pm$ standard deviation over 3 seeds. Inversion replaces the targeted
  labels with guaranteed wrong ones. Bold marks the better method per cell.
  The first four tasks are within the pool's reach; race is a task the pool
  covers poorly, where the constraint caps performance below LoRA even on clean
  data, though $z$ stays flat under corruption while LoRA does not.}
  \label{tab:inversion}
  \small
  \begin{tabular}{lcccccc}
    \toprule
    & \multicolumn{2}{c}{clean} & \multicolumn{2}{c}{50\% inverted}
    & \multicolumn{2}{c}{100\% inverted} \\
    \cmidrule(lr){2-3}\cmidrule(lr){4-5}\cmidrule(lr){6-7}
    Task & LoRA & $z$ & LoRA & $z$ & LoRA & $z$ \\
    \midrule
    wiki\_qa      & 0.99 & 0.99 & 0.49 & \textbf{0.99} & 0.03 & \textbf{0.92} \\
    qasc          & 0.99 & 0.98 & 0.59 & \textbf{0.92} & 0.03 & \textbf{0.92} \\
    amazon        & 0.94 & 0.95 & 0.58 & \textbf{0.95} & 0.05 & \textbf{0.96} \\
    social\_iqa   & 0.72 & 0.68 & 0.45 & \textbf{0.66} & 0.26 & \textbf{0.62} \\
    \midrule
    race (weak pool) & \textbf{0.31} & 0.13 & \textbf{0.29} & 0.17 & \textbf{0.22} & 0.16 \\
    \bottomrule
  \end{tabular}
\end{table}

\subsection{Incompatibility signal (a pool-relative OOD detector)}
\begin{figure}[t]
  \centering
  \includegraphics[width=0.55\linewidth]{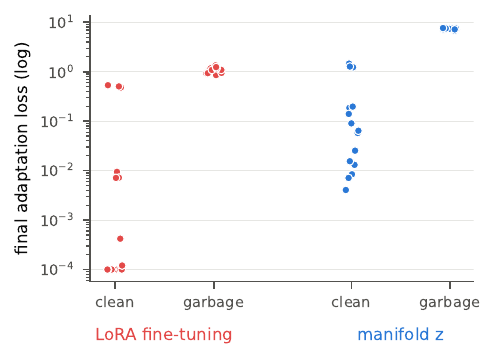}
  \caption{Final adaptation loss on clean versus garbage targets (log scale,
  one point per task and seed). Constrained adaptation cannot descend on
  garbage (median 7.4 versus 0.06 clean, a $120\times$ margin); LoRA also
  separates but by a much smaller margin (1.0 versus 0.0) and, crucially,
  ships a corrupted model.}
  \label{fig:ood}
\end{figure}
The same mechanism that blocks the poison produces a detector: data the pool
cannot express keeps the adaptation loss high. Thresholding the final
adaptation loss separates clean from garbage with AUROC 1.0 for both methods on
these samples, so detectability itself is not unique to our method. Two things
are: the margin is two orders of magnitude for constrained adaptation versus
one for LoRA, and for constrained adaptation a high loss \emph{coincides} with
safety (the poison was not learned), whereas LoRA reaches a low loss precisely
by learning the poison. The detector is relative to the pool, which is a
genuine caveat: a legitimate task far from everything in the pool would also
plateau, a false positive. Sensitivity to poison and coverage of
hard-but-legitimate tasks are two sides of the pool's span; we therefore read
the plateau as ``outside the pool'' rather than ``adversarial'' per se.

\subsection{Sequential adaptation}
\label{sec:sequential}
\begin{figure}[t]
  \centering
  \includegraphics[width=0.55\linewidth]{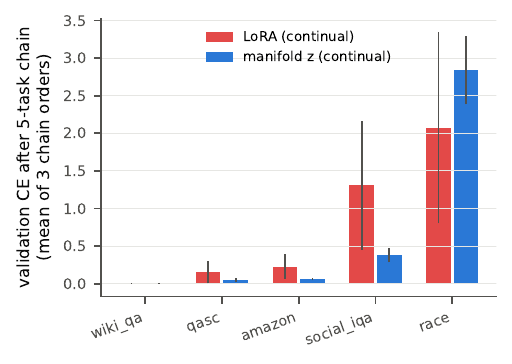}
  \caption{Validation CE on all five tasks after learning them in sequence with
  a single continually-updated adapter, mean over the three chain orders. On the
  tasks the pool covers, $z$ forgets less than LoRA; race is the exception.}
  \label{fig:sequential}
\end{figure}
We train one adapter continually over a chain of five tasks (three seeds, three
chain orders) and measure both forgetting, the rise in a task's validation CE
between just after it was learned and the end of the chain, and recovery,
re-learning the first task from 16 examples after the full chain. Averaged over
seeds, constrained adaptation forgets slightly less than LoRA ($0.40$ versus
$0.47$ mean CE rise) and far more consistently ($\pm 0.11$ versus $\pm 0.41$):
LoRA forgets catastrophically on some chains (one task's CE rising above 2.5),
while the constraint bounds the damage. On the pool-covered tasks the gap is
clear (Figure~\ref{fig:sequential}); race, again, is where the constraint hurts.
Recovery is mixed and dominated by whichever task is first in the chain: it
favors $z$ when that task is pool-covered (0.06 versus 0.20 CE when amazon is
first) and LoRA when it is race. We report the aggregate without overclaiming a
recovery advantage.

\subsection{Nonlinear generator versus PCA}
\label{sec:generator}
\begin{figure}[t]
  \centering
  \includegraphics[width=0.55\linewidth]{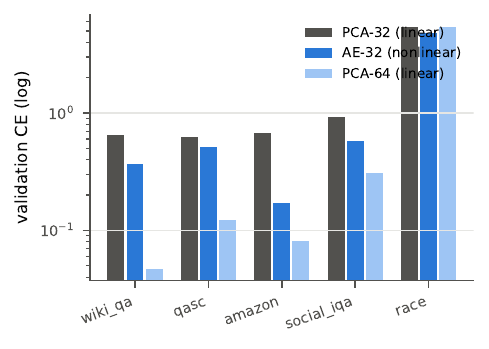}
  \caption{Reconstructing a held-out adapter through a code of fixed size, then
  evaluating its function. A nonlinear autoencoder at bottleneck 32 (AE-32)
  beats a linear PCA truncation at 32 components on every task, but a linear
  PCA at 64 components beats them both.}
  \label{fig:generator}
\end{figure}
Is the affine subspace of Section~\ref{sec:method} leaving anything on the
table? We compare, at equal code size, a linear PCA truncation and a small
autoencoder trained on the other 195 adapters' coordinates, reconstructing each
held-out adapter and scoring its function. At bottleneck 32 the autoencoder
beats PCA-32 on all five tasks (Figure~\ref{fig:generator}), so the pool does
have mild nonlinear structure that a curved generator captures. But PCA-64, a
larger linear subspace, beats the nonlinear AE-32: nonlinearity buys code
efficiency, not peak fidelity. We therefore keep the linear subspace as the
main method for its simplicity and report the autoencoder as evidence that
``manifold'' is more than a figure of speech, without claiming the nonlinear
generator is necessary.

\subsection{Controls: leakage and low-dimension}
\label{sec:controls}
Two controls address the most natural objections. \textbf{Dataset-family
holdout.} Our pool contains several adapters per source dataset (nine
amazon\_polarity prompts, nine wiki\_qa, and so on), so leaving out only the
target adapter still leaves near-siblings in the basis. We rebuild the basis
excluding \emph{every} adapter from the target's dataset (4 to 9 removed per
task) and re-run clean and 100\% inversion. \textbf{Random subspace.} To
separate the pool's semantics from the mere fact of a 128-dimensional
constraint, we replace the pool with 128 random rank-16 pseudo-adapters
(Gaussian, Frobenius-norm-matched) and adapt in their span.
Both controls come out as the mechanism predicts (Table~\ref{tab:controls}).
Removing every same-dataset sibling barely changes constrained adaptation: at
100\% inversion it keeps 0.65 to 0.99 exact match (versus 0.62 to 0.96 with the
full pool), while LoRA still collapses to 0.03 to 0.26. The robustness is
therefore not a leakage artifact. The random subspace, by contrast, is useless:
at equal size and budget it reaches only 0.06 to 0.53 exact match on
\emph{clean} data (versus 0.68 to 0.99 for the pool), so a 128-dimensional
constraint alone buys neither utility nor meaningful robustness. What the pool
provides, that a random subspace of the same size does not, is the combination
of clean utility and poison resistance; low dimension alone gives neither.

A third control asks whether the protection is just slower learning. We run
LoRA with strong regularization (weight decay 0.1 plus dropout 0.3, and weight
decay 1.0). Both keep clean performance (0.94 to 0.99) and both still collapse
under inversion (0.01 to 0.05 exact match on the first three tasks), identical
to plain LoRA. A norm penalty does not restrict \emph{which} solutions are
reachable, only how fast they are reached, so it offers no poison resistance.
The constraint is doing something a regularizer cannot.

A fourth control addresses an asymmetry in the comparison: constrained
adaptation starts from $z=0$, which decodes to the pool-mean adapter, a
non-trivial prior, whereas LoRA starts from a near-zero update. One might worry
that the poison resistance is just the method staying near a good starting
point. It is not: on the four pool-covered tasks the frozen mean adapter alone
(no training) reaches only 0.19 to 0.53 exact match on clean data, while
adaptation in the subspace reaches 0.68 to 0.99 (for example wiki\_qa 0.44 to
0.99, social\_iqa 0.19 to 0.68). The method learns substantially beyond its
prior; the prior is part of the method, and we report its standalone score so
the improvement is visible. On race, where the pool is weak, the mean adapter
and the adapted model are both near the base (0.11 and 0.13), consistent with
the plasticity ceiling.

\begin{table}[t]
  \centering
  \caption{Controls, exact match, mean over 3 seeds. Family holdout removes all
  4 to 9 same-dataset adapters from the basis. Random subspace replaces the pool
  with 128 norm-matched random rank-16 directions.}
  \label{tab:controls}
  \small
  \begin{tabular}{lcc c cc}
    \toprule
    & \multicolumn{2}{c}{family holdout} & & \multicolumn{2}{c}{random subspace} \\
    \cmidrule(lr){2-3}\cmidrule(lr){5-6}
    Task & $z$ clean & $z$ flip100 & & clean & flip100 \\
    \midrule
    wiki\_qa    & 0.99 & 0.99 & & 0.15 & 0.03 \\
    qasc        & 0.98 & 0.94 & & 0.29 & 0.09 \\
    amazon      & 0.95 & 0.96 & & 0.53 & 0.38 \\
    social\_iqa & 0.70 & 0.65 & & 0.06 & 0.02 \\
    \bottomrule
  \end{tabular}
\end{table}

\subsection{Adaptive backdoor attack}
\label{sec:adaptive}
\begin{figure}[t]
  \centering
  \includegraphics[width=0.55\linewidth]{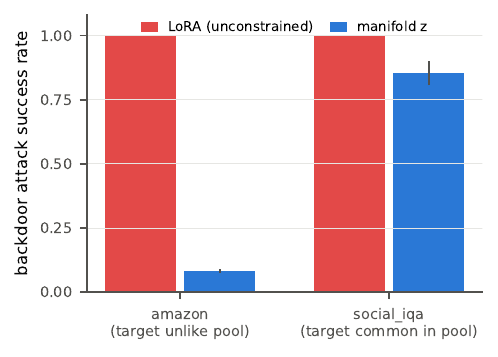}
  \caption{Adaptive backdoor. An attacker inserts a rare trigger token and
  optimizes (in the maxASR mode, ignoring clean accuracy) to make triggered
  inputs output a fixed target. Unconstrained LoRA reaches 100\% attack success
  on both tasks. Constrained adaptation caps the attack at 8\% when the target
  behavior is unlike anything in the pool (amazon: force ``negative''
  regardless of sentiment) but only at 85\% when the target coincides with a
  behavior the pool already exhibits (social\_iqa, a yes/no validity task where
  ``always No'' is a nearby legitimate direction).}
  \label{fig:adaptive}
\end{figure}
The attacks so far fail because they lie outside the subspace. The sharper test
is an \emph{adaptive} attacker who searches for a harmful behavior expressible
\emph{inside} it: a backdoor that maps a trigger token to a target output while
preserving clean accuracy. We give the attacker the defense itself, optimizing
the code $z$ (and, as baseline, LoRA) directly on triggered data, and we exclude
the target's dataset family from the basis.

The result is not uniform, and the non-uniformity is the finding. Unconstrained
LoRA plants the backdoor at will (attack success 0.84 to 1.00 while keeping
clean accuracy, 1.00 when pushed maximally). Constrained adaptation blocks it on
amazon (ceiling 0.08) but not on social\_iqa (ceiling 0.85). The difference is
consistent with the same principle as the rest of the paper: the backdoor
appears blocked when its target behavior lies outside the pool's common
directions. On amazon the target ``output negative regardless of sentiment''
contradicts every sentiment adapter, so no $z$ we found expresses it; on
social\_iqa the task is itself a yes/no validity judgment, ``always No'' is a
frequent pool direction, and the trigger latches onto it. With two tasks we
cannot claim a general law, only a plausible and testable one: the security
boundary looks pool-relative, and its strength may be anticipated from whether
the feared behavior is representable in the pool. This moves a binary claim
(``it blocks backdoors'') toward a characterized one (``it blocks backdoors whose
target is not already a pool behavior''), which is the more defensible and more
useful statement.

\section{Limitations and threats to validity}
\label{sec:limits}
\textbf{Clean pool assumed.} The central assumption is that the adapter pool is
trusted. We secure the \emph{adaptation data}, not the pool: an attacker who can
inject a malicious adapter into the pool moves the subspace and defeats the
protection. Securing the pool is a separate supply-chain problem (provenance,
vetting, anomaly detection on the pool spectrum) and is out of scope.
\textbf{No formal guarantee.} The subspace is spanned by legitimate adapters but
is not their convex hull; it contains far extrapolations, negative
combinations, and behaviors present in no single adapter, and composed adapters
can be unsafe even when each is benign~\citep{composition2026}. Our results are
empirical protection against the tested attacks, not a proof of behavioral
safety. A convex-hull restriction ($g_j \ge 0, \sum_j g_j = 1$) or a
trust-region ($z^\top \Sigma^{-1} z \le \rho$) would bound extrapolation and is
future work.
\textbf{Threat-model breadth.} We test label inversion, garbage targets, and an
adaptive trigger backdoor (Section~\ref{sec:adaptive}). The adaptive result
bounds the claim honestly: the subspace is a security boundary only against
behaviors it does not already contain, and a determined attacker whose target
aligns with the pool (social\_iqa) partly succeeds. Remaining stronger cases:
low-rate backdoors (0.5--5\%) that preserve clean
accuracy~\citep{backdoor2024}, clean-label poisoning, and multi-token or
optimized triggers.
\textbf{Plasticity ceiling.} Performance is capped by pool coverage; on
poorly-covered tasks (race) unconstrained fine-tuning wins on clean data.
\textbf{Scope and baselines.} Single base model, single pool, English P3 tasks,
independently trained adapters. Section~\ref{sec:controls} passes four controls
(dataset-family holdout, random subspace, strong-regularization LoRA, and a
frozen prior baseline). A DoRA variant and EigenLoRAx as a direct comparison
remain to be run.

\section{Conclusion}
Restricting adaptation to a subspace estimated from trusted adapters turns three
empirical safety properties into consequences of geometry: the poisoned
objectives we test fail to be represented, garbage data announces itself in the
loss, and few-shot adaptation is regularized by construction. Concretely, this
is the difference between an on-device assistant that can only become a variant
of a valid assistant and one that can be steered anywhere, and between a
training pipeline that halts on an anomalous-loss spike and one that ships the
poison silently. The protection is empirical and bounded, and the adaptive backdoor sharpens the
picture: across our tasks the subspace stops an attack when the attack's target
behavior is not already something the pool can do, and lets it through when it
is. This is not a weakness of the analysis but its content, since whether a
feared behavior is in the pool is something one can check before deployment.
What we establish, on the tasks we test, is that a trusted adapter subspace is a
restricted-adaptation policy whose poisoning resistance is measurable, tracks
pool coverage, and comes with a built-in incompatibility signal.

\section{Reproducibility}
All scripts, seeds, and incremental JSON outputs are in the repository. Every
experiment runs on one 12 GB GPU in bf16; the spectrum analysis runs on CPU.

\bibliographystyle{plainnat}
\bibliography{references}

@inproceedings{rusu2019leo,
  title     = {Meta-Learning with Latent Embedding Optimization},
  author    = {Rusu, Andrei A. and Rao, Dushyant and Sygnowski, Jakub and
               Vinyals, Oriol and Pascanu, Razvan and Osindero, Simon and
               Hadsell, Raia},
  booktitle = {International Conference on Learning Representations},
  year      = {2019}
}

@inproceedings{charakorn2025t2l,
  title     = {Text-to-LoRA: Instant Transformer Adaption},
  author    = {Charakorn, Rujikorn and Cetin, Edoardo and Tang, Yujin and
               Lange, Robert Tjarko},
  booktitle = {International Conference on Machine Learning},
  year      = {2025},
  note      = {arXiv:2506.06105}
}

@article{dnd2025,
  title   = {Drag-and-Drop LLMs: Zero-Shot Prompt-to-Weights},
  author  = {Liang, Zhiyuan and Tang, Dongwen and Zhou, Yuhao and Zhao, Xuanlei
             and Shi, Mingjia and Zhao, Wangbo and Li, Zekai and Wang, Peihao
             and Sch{\"u}rholt, Konstantin and Borth, Damian and Bronstein,
             Michael M. and You, Yang and Wang, Zhangyang and Wang, Kai},
  journal = {arXiv preprint arXiv:2506.16406},
  year    = {2025}
}

@article{cts2024,
  title   = {Compress then Serve: Serving Thousands of LoRA Adapters with
             Little Overhead},
  author  = {Br{\"u}el-Gabrielsson, Rickard and Zhu, Jiacheng and Bhardwaj,
             Onkar and Choshen, Leshem and Greenewald, Kristjan and
             Yurochkin, Mikhail and Solomon, Justin},
  journal = {arXiv preprint arXiv:2407.00066},
  year    = {2024}
}

@inproceedings{lorahub2024,
  title     = {LoraHub: Efficient Cross-Task Generalization via Dynamic LoRA
               Composition},
  author    = {Huang, Chengsong and Liu, Qian and Lin, Bill Yuchen and Pang,
               Tianyu and Du, Chao and Lin, Min},
  booktitle = {Conference on Language Modeling},
  year      = {2024}
}

@inproceedings{schurholt2022,
  title     = {Model Zoos: A Dataset of Diverse Populations of Neural Network
               Models},
  author    = {Sch{\"u}rholt, Konstantin and Taskiran, Diyar and Knyazev,
               Boris and Gir{\'o}-i-Nieto, Xavier and Borth, Damian},
  booktitle = {Advances in Neural Information Processing Systems, Datasets and
               Benchmarks Track},
  year      = {2022}
}

@inproceedings{gitrebasin2023,
  title     = {Git Re-Basin: Merging Models modulo Permutation Symmetries},
  author    = {Ainsworth, Samuel K. and Hayase, Jonathan and Srinivasa,
               Siddhartha},
  booktitle = {International Conference on Learning Representations},
  year      = {2023}
}

@inproceedings{wan2023poisoning,
  title     = {Poisoning Language Models During Instruction Tuning},
  author    = {Wan, Alexander and Wallace, Eric and Shen, Sheng and Klein, Dan},
  booktitle = {International Conference on Machine Learning},
  year      = {2023}
}

@inproceedings{qi2024finetuning,
  title     = {Fine-tuning Aligned Language Models Compromises Safety, Even
               When Users Do Not Intend To!},
  author    = {Qi, Xiangyu and Zeng, Yi and Xie, Tinghao and Chen, Pin-Yu and
               Jia, Ruoxi and Mittal, Prateek and Henderson, Peter},
  booktitle = {International Conference on Learning Representations},
  year      = {2024}
}

@article{eigenlorax2025,
  title   = {EigenLoRAx: Recycling Adapters to Find Principal Subspaces for
             Resource-Efficient Adaptation and Inference},
  author  = {Kaushik, Prakhar and Vaidya, Ankit and Chaudhari, Shravan and
             Yuille, Alan},
  journal = {arXiv preprint arXiv:2502.04700},
  year    = {2025}
}

@article{composition2026,
  title   = {Colluding LoRA: A Compositional Vulnerability in LLM Safety
             Alignment},
  author  = {Ding, Sihao and others},
  journal = {arXiv preprint arXiv:2603.12681},
  year    = {2026}
}

@article{backdoor2024,
  title   = {A Study of Backdoors in Instruction Fine-tuned Language Models},
  author  = {Raghuram, Jayaram and Kesidis, George and Miller, David J.},
  journal = {arXiv preprint arXiv:2406.07778},
  year    = {2024}
}
\end{document}